%% file: influenceSamplingArxiv.tex
\documentclass{article}

\usepackage[utf8]{inputenc} %
\usepackage[T1]{fontenc}    %
\usepackage{hyperref}       %
\usepackage{url}            %
\usepackage{amsfonts}       %
\usepackage{nicefrac}       %

\usepackage{graphicx}
\usepackage{amsmath,amssymb,amsthm,bm,enumerate}
\usepackage{latexsym,color,verbatim,minipage-marginpar,caption,multirow}
\usepackage{multicol}

\usepackage{array} %
\usepackage{paralist} %
\usepackage{amsmath}
\usepackage{amsfonts}
\usepackage{verbatim} %
\usepackage{subfig} %
\usepackage{algorithm}
\usepackage{algpseudocode}

\usepackage{todonotes}

\input{macros.tex}

\title{Optimal Sub-sampling with Influence Functions}
\author{
  Daniel~Ting \\
  Tableau Software \\
  Seattle, WA 98105 \\
  \texttt{dting@tableau.com} \\
  \and
  Eric~Brochu \\
  Tableau Software \\
  Seattle, WA 98105 \\
  \texttt{ebrochu@tableau.com}
}

\begin{document}

\maketitle

\begin{abstract}
Sub-sampling is a common and often effective method to deal with the computational challenges of large datasets. 
However, for most statistical models, there is no well-motivated approach for drawing a non-uniform subsample. We show that the concept of an asymptotically linear estimator 
and the associated influence function leads to optimal sampling procedures
for a wide class of popular models. Furthermore, for linear regression models which have well-studied procedures for non-uniform sub-sampling, we show our optimal influence function based method outperforms previous approaches. We empirically show the improved performance of our method on real datasets. 
\end{abstract}

\section{Introduction}

As the amount of data in the world increases, the question arises as to how best to deal with the large datasets. 
The associated tasks can be varied as they include training models on big data, rapid prototyping when datasets do not fit on a single machine, parameter tuning and model selection,
and
data exploration and visualization of ``important'' data points such as outliers.
Although a number of data sketching techniques such as random projections exist to reduce the size of the data and the associated computational costs, many can only be applied to some pre-specified task and may require custom code to use. However, sampling methods are a common method for dealing with the problem of size as they provide an exceptionally flexible summarization of the data that can be applied to almost all tasks in a simple, straightforward manner. 

The simplest sampling method is uniform random sampling. 
However, it is inefficient as it does not exploit any notion of the importance of a data point for the relevant tasks.
We would ideally like to sample the data \emph{efficiently}, preferentially sampling the data that will accurately approximate the estimates from the full data set while avoiding wasting resources on data that are nearly irrelevant. 
Although several approaches introduce preferential sampling probabilities such as sampling based on leverage scores \cite{drineas2012fast} or gradients \cite{zhu2016gradient},
we show that the resulting sampling probabilities are still inefficient and can demonstrate pathological behavior. 
Furthermore, most methods are derived only for a linear regression model.
To the authors' knowledge, there is limited work on sampling for more general machine learning models outside of logistic regression which was studied by \cite{fithian2014local}.
In this paper, we examine the problem of finding optimal sub-sampling probabilities for nearly any estimation problem.

To this end, we propose using the \emph{influence function} as a measure of sampling importance.
The influence function measures the change in the objective or values of interest due to a single point.
It is a particularly general approach as many model and estimators, such as maximum likelihood and M-estimators, can be cast in the framework and can work with non-differentiable objectives.
We prove that the regularized version of our sampling design is asymptotically optimal among all regularized designs of the same expected size. 
This is a substantially stronger result than other results that minimize a loose probabilistic upper bound on the error.

Beyond the improved performance of our method, the influence-based approach allows one to fundamentally understand the problem of optimal sub-sampling. Rather than proposing an ad hoc method and analyzing its theoretical properties,
the influence function and the notion of asymptotically linear estimators 
reduces the problem of accurately approximating the estimate from the full data set to the problem of calculating the mean of influence functions. Thus, the problem of finding an optimal sampling design or probabilities for estimating a model can be converted to the more straightforward problem of optimal sampling design for a mean.
This design can depend on the specific task on hand. In particular, good sampling designs to estimate the parameters of a model can substantially differ from good sampling designs 
to optimize the resulting predictions even when the same model is used in both. This fact is also borne out in our experimental results.
We explicitly derive sampling probabilities for linear regression, quantile regression, and generalized linear models. In doing so, we are able to separate 
the ``influence'' of the residuals $y_i - \hat{y}_i$ from the ``influence'' of the regression design or predictors $X$. As a result, we are able to show that
existing approaches often only appropriately exploits one of the two, whereas our method appropriately incorporates both.

\vspace{-0.2cm}
\subsection{Related work}
\vspace{-0.1cm}
A number of methods exist for sub-sampling when the relevant task is linear $L_p$ regression or matrix approximation. 
The dominant approach in the literature for least squares regression is based on statistical leverage scores \cite{ma2015statistical}, \cite{drineas2012fast}.
A number of papers \cite{clarkson2005subgradient}, \cite{meng2013low}, \cite{dasgupta2009sampling}, \cite{clarkson2016fast} address more general $L_p$ linear regression problems
and derive a corresponding leverage for $L_p$ regression.
These methods focus on generating sampling designs from the design matrix or predictors $X$, and make no or limited use of the responses $Y$.
The resulting sampling designs are obtained via a relatively expensive to compute and complex random projection or low distortion embedding.

For linear regression models that make use of the responses, the gradient-based approach of \cite{zhu2016gradient}  and the Uluru algorithm in \cite{dhillon2013new} provide methods that subsample the data 
based on residuals given a pilot estimate of the coefficients. 
Although the results for the gradient-based approach only deal with linear least squares regression, the 
gradient can be computed for any differentiable loss function. 
We show that the Uluru algorithm is a special case of our optimal sampling procedure when applied to the coefficients of a linear regression problem.
Interestingly, although the theoretical results for Uluru deal are stated in terms of the prediction error, we find Uluru is suboptimal in this regime. Outside of linear models, local case-control sampling \cite{fithian2014local} provides an effective sampling method for logistic regression based on the residuals as well.

A number of other other techniques are relevant for fast model fitting on large datasets. 
Stochastic gradient descent (SGD) employs sampling and is useful in large-scale learning problems \cite{bottou2010large}. 
Other techniques such as \cite{meng2014lsrn} use random projections to speed up the computation in the inner loop of a model fitting algorithm. 
In these cases, the data size is not reduced and the result is an estimate for the specific model being fit.
In contrast, we focus on sampling as a way to reduce computational complexity for a specific model or models 
while also being able to further use the sample for other purposes, for example, in parameter tuning.

\vspace{-0.2cm}
\section{Influence functions and asymptotically linear estimators}
\vspace{-0.1cm}
\label{sec:asymptotically linear}
One key idea in this paper is that many parameter estimators can be asymptotically expressed as a mean of influence vectors. This allows us to express the problem of optimal sub-sampling for a statistical model in terms of optimal sub-sampling for a mean. We give a brief overview of the theoretical machinery needed for our method and refer the interested reader to \cite{vandervaart} for more details.

We consider the class of estimators are plug-in estimators
$\hat{\theta}(P)$ taking a distribution to a real-valued vector of parameter estimates. This is a highly flexible class of estimators. For example, any M-estimator $\hat{\theta}(P) = \arg\!\min_t \E_{P} Loss(t, X)$ is of this form. The argument for the M-estimator is  the empirical distribution $\mathbb{P}_n$.
 
When it exists, the influence function for this estimator is defined by its Gateaux derivative
\begin{align}
\psi_P(x) &= \lim_{\epsilon \to 0^+}  \frac{1}{\epsilon}\left(\hat{\theta}\left((1-\epsilon)P + \epsilon \delta_x\right) - \hat{\theta}(P)\right)
\end{align}
where $\delta_x$ is the Dirac delta measure at $x$. 
It represents the infinitesimal change in the estimate by adding the point $x$ to the sample. 
The estimator $\hat{\theta}$ is an {\em asymptotically linear estimator} 
with influence function $\psi$ if it satisfies 
\vspace{-0.2cm}
\begin{align}
\label{eqn:asymptotically linear}
\sqrt{n}\left(\hat{\theta} - \theta\right) = 
\frac{1}{\sqrt{n}} \sum_{i=1}^n \psi(X_i) + o_p(1),
\end{align}
with $\E \psi = 0$ and $\E \psi^T \psi < \infty$.
$o_p(1)$ denotes convergence in probability in some normed space.

Asymptotically linear estimators are pervasive in statistical modeling. Under sufficient regularity conditions, the previously mentioned M-estimators and maximum likelihood estimators as well as Z-estimators, non-degenerate U-statistics, and Generalized Method of Moments estimators are asymptotically linear estimators.
For maximum likelihood estimation with correctly specified  and sufficiently regular models, the influence function can be described in terms of the derivative of the log-likelihood with respect to the parameter, in other words, the score function
$s_\theta(x)$. The influence function can be related to score by 
\begin{align}
\label{eqn:mle influence}
s_{\theta}(x) &= \partial \ell(\theta; x) / \partial \theta, \qquad \qquad
\psi_{\theta} = I_{\theta}^{-1} s_{\theta}
\end{align}
where $I_\theta$ is the Fisher information.

Unlike gradient-based sampling, there is no differentiability requirement. Thus, non-differentiable likelihoods such as a double exponential location family or quantile regressions fit in the framework. Furthermore, we find both in theory and practice that gradient-based sampling accounts for the regression design (i.e. the predictors) on an inappropriate scale. As a consequence, even sampling using only the residuals as weights and completely ignoring the regression design often performs better than the full gradient
as shown in figure \ref{fig:CASP}.

As the influence function asymptotically encodes the effect of a single data point, it is sensible to use it to determine the point's sampling weight. The asymptotic form of an asymptotically linear estimator shows that the problem of sampling for the estimator can be reduced to a problem of estimating a multivariate mean.

\subsection{Linear Least Squares Influence}
In the context of linear regression, the influence function for the coefficients $\hat{\theta}$ is given by 
\begin{align}
\label{eqn:linear influence function}
\psi_{P_\theta}(x_i, y_i) &= (y_i - x_i^T \theta) \Sigma^{-1} x_i
\end{align}
where $\Sigma = \frac{1}{n}\left(X^T X\right)$ is the empirical second moment matrix and $\theta$ are the true coefficients. 

Taking the norm of the influence yields a sampling weight that differs from gradient-based sampling only by the scaling. The weight is proportional to  $\|(y_i - \hat{y}_i)\Sigma^{-1} x_i\|$ rather than $\|(y_i - \hat{y}_i)x_i\|$. 
The influence-based scaling is the more sensible of the two.  
For example, if one wishes to make the $j^{th}$ coordinate more important,  $\theta_j$ must be scaled to be larger and $x_{ij}$ to be smaller.
Under gradient-based sampling, a previously important point with a large $x_{ij}$ is perversely made less important as $x_{ij}$ is scaled downwards.
Influence-based sampling correctly increases the importance of the point.

\subsection{Influence on Predictions}
\label{sec:prediction influence}
The notion of influence can be extended beyond estimators for parameters. For example,
points can be sampled according to their influence on predictions rather than on the coefficients themselves. For many prediction problems, this measure may be more sensible as one neither cares about the exact values of the coefficient nor the scale of the variables if the predictive performance is good.

The influence on the prediction is easily derived from the influence on the coefficients. When each prediction $\hat{y}_i(\theta)$ is a twice differentiable function of the parameter $\theta$,
then, with a slight abuse of notation, 
the influence on the vector of predictions is given by the chain rule
\begin{align}
\psi^{(pred)}_{P_\theta}(x_i, y_i) = \hat{y}'(\theta) \frac{d \theta}{d \delta_{(x_i,y_i)}} =
\hat{y}'(\theta) \psi_{P_\theta}(x_i, y_i).
\end{align}
In the context of linear least squares regression, the influence is 
$\psi^{(pred)}_{P_\theta}(x_i, y_i) = X (X^TX)^{-1} x_i (y_i - \hat{y}_i) = r_i H_{i\cdot}$
where $r_i$ is the residual error and $H_{\cdot i}$ is the $i^{th}$ column of the hat matrix.

Since the hat matrix $H$ is idempotent and symmetric, it follows that the squared norm 
$\| H_{i \cdot} \|^2 = e_i^T  H^T H e_i = e_i^T H e_i = H_{ii}$. 
A sensible univariate measure for a sampling weight is thus $|r_i| \sqrt{H_{ii}}$. 

The prediction on the influence provides a strong connection to leverage-based sampling. In the classical statistical setting where the experimenter does not have knowledge of the response $Y_i$ when setting the regression design $X$, a sensible measure of influence takes the expectation over the unknown $Y_i$. In this case, one obtains a sampling weight proportional to the root leverage score $\sqrt{H_{ii}}$.
If only the influence on the prediction $\hat{y}_i$ is considered,
the influence is $\psi^{(pred,i)}_{P_\theta}(x_i, y_i) = x_i (X^TX)^{-1} x_i (y_i - \hat{y}_i) = r_i H_{ii}$. Taking an expectation over an unknown $Y_i$ exactly gives back leverage score sampling.
Thus, we see that leverage-based sampling throws away two pieces of information that are available in sampling for computational problem, the effect of the regression design on points other than the $i^{th}$ point and more importantly, the information on the response $Y_i$.

\section{Sampling design and Estimation from samples}
Reducing computational costs using the influence function requires three components:
1) a method to estimate the influence function, 2) a method to convert the influence function into a sampling design and sampling probabilities, and 3) a method to provide good estimates from the sample. 
We address these in reverse order.

\subsection{Estimation from samples}
A good sub-sampling procedure allows one to accurately approximate the estimator $\hat{\theta}$. This is a somewhat different problem from approximating the true parameters.
Let $\tilde{\theta}$ be an estimator based on a subsample. As $\tilde{\theta}$ cannot be expected to improve upon the estimator on the full data $\hat{\theta}(\mathbb{P}_n)$, it is reasonable to assume that its expectation is close to or equal to $\hat{\theta}(\mathbb{P}_n)$, and
the variance $\var(\tilde{\theta}) \approx \E \var(\tilde{\theta} | \hat{\theta}- \theta)$.
In other words, the subsample should be the best sample for approximating the 
deviation $\hat{\theta} - \theta$.
This deviation  has the asymptotically linear form given in equation \ref{eqn:asymptotically linear}, so that approximating the deviation is asymptotically equivalent to approximating the mean of influences.

The quantities of interest such as the objective or estimators that we consider in this paper can be expressed as functions of the empirical distribution $\P$.
For any subsampling procedure, an unbiased estimate of the true empirical distribution can be obtained by reweighting the sample points by their inverse sampling probability. 
In the case of mean estimation, the resulting estimator called the Horvitz-Thompson estimator $\hat{\mu} = n^{-1} \sum_i X_i Z_i / \pi_i$ where $Z_i$ is the indicator that a point was sampled and $\pi_i = \E Z_i$ is the sampling probability.
This can also be applied to obtain an unbiased estimate $\hat{\mathbb{P}}$ of the true empirical distribution and resulting parameter estimator
\begin{align}
\hat{\mathbb{P}} &= \left(\sum_{i=1}^n \frac{1}{\pi_i}\right)^{-1}\sum_{i=1}^n \frac{Z_i}{\pi_i} \delta_{x_i}, \qquad \qquad
\tilde{\theta} = \hat{\theta}(\hat{\mathbb{P}}) .
\end{align}

\subsection{Optimal design}
\label{sec:optimal design}
The next question is how to choose an appropriate sampling procedure
and convert a vector valued influence to a univariate sampling probability. 
A sample can then be drawn by independently selecting each item according to its sampling probability. This scheme is called Poisson sampling.
Oftentimes, there is a sampling objective that should be minimized as well a size budget for the number of samples that can be stored.

Consider the objective of minimizing $L_2$ error of the parameters, and suppose
the original data points are drawn i.i.d.\@ from some distribution.
One has $\|\hat{\theta} - \theta\|^2 \approx \frac{1}{n^2} \sum_i \|\psi_i\|^2 \propto Trace(\var\, \psi_i)$.
If each $\psi_i$ is sampled independently with probability $\pi_i$, the optimal probabilities $\pi_i$ for drawing a sample of expected size $m$ 
minimizes the objective $\min V\left( \sum_i \|\psi_i\| Z_i / \pi_i\right)$
subject to the constraints $\sum_i \pi_i = m$ and $\pi_i \leq 1$. Applying the method of 
Lagrange multipliers gives $\pi_i \propto \|\psi_i\|$ for all $\pi_i \neq 1$.
This gives an instance of probability proportional to size (PPS) sampling where the auxiliary measure of size is the norm of the influence function for the parameters. 
Similarly if the objective is minimizing the $L_2$ error in the predictions,
an appropriate sampling scheme uses PPS sampling with size equal to the norm of the influence function of the predictions.

We note that the problem of choosing a sampling weight is non-obvious. For least squares regression,
the optimal choice for approximating the loss at $\hat{\theta}$ is to sample with probability proportional
to the squared residual $(y_i - \hat{y}_i)^2$. We found this to be a poor choice for approximating the coefficients in our experiments. %
 The reason
is that the raw value of the loss is unimportant. Asymptotically, only the gradient and Hessian are relevant.
We also note that although past work \cite{drineas2012fast}  sampled with probability proportional to the leverage score, our work surprisingly shows that the square root of the leverage may be a more appropriate measure.

In other cases, one may be interested in quantities other than the predictions or all the coefficients together. For example, if one wishes to study the effect of gas and electricity pricing on consumption, any reasonable model would adjust for weather effects. In this case, only a subset of coefficients may be of interest while others, like those for weather, are nuisance parameters. 
In this case, one may use the influence restricted to the coefficients of interest.

\subsection{Regularization}
Since each sampled point is weighted by its inverse sampling probability,
the resulting estimate may have high variance if a sampling probability is too small. 
The solution in this paper is to add a small amount of regularization
to ensure that no sampling probability is too small. 
This ensures convergence $\gamma_n(\hat{\mathbb{P}} -\mathbb{P}_n) \convd T$ to some limit process $T$ under an appropriate scaling $\gamma_n$.
In deriving an optimal sampling design, 
this corresponds to adding the convex constraint that $\alpha \leq \pi_i \leq 1$ to the optimization in \ref{sec:optimal design}.
The resulting sampling probabilities have 
$\pi_i \propto \| \psi(x_i) \|$ if $\pi_i \neq 1$ or $\alpha$.

\subsection{Influence function estimation}
Thus far, we have derived the exact influence function given a population distribution or
in finite sample cases, the empirical distribution.  In most cases,
the influence function depends on the true parameter $\theta$. 
A simple estimate of the influence function uses a pilot estimate $\theta_0$ as a substitute.
The pilot estimate may be readily available or easily obtained.
For example, in many machine learning applications, the estimated parameters from one day may be used as a pilot estimate for the next day. If one has no pre-existing pilot estimate, then one can first draw a uniform sample, or even a reasonable convenience sample, from the data
to form a pilot that can be used on the remaining data. 

A second approach requires only the ability to compute the gradient of the log-likelihood, in other words the score function. As described in section \ref{sec:asymptotically linear}, for a sufficiently regular family, the influence function for the maximum likelihood estimate is a rescaling of the score function $s$ by the inverse of the Fisher information $I_\theta = \E ss^T$. 
A simple procedure to estimate the influence function is to 1) start with a pilot estimate $\theta_0$, 2) compute an online estimate of the covariance $\hat{V}_{\theta_0}$ for the scores at $\theta_0$, and 3) rescale the score $\hat{\psi}(x) = \hat{V}_{\theta_0}^{-1} s_{\theta_0}(x)$.

In experiments we find that the approximation to the leverage score or non-residual component of the influence can negatively affect the quality of the sample. In such cases, approximating the component with a constant yields the extraordinarily simple---but often effective---sampling design with probability proportional to the residual.

\subsection{Influence function computation}
For many estimators, computing the influence function requires a matrix inversion or pseudo-inverse. This may be costly to compute. 
A simple strategy is to replace $V$ with its diagonal, in other words, the Jacobi preconditioner. 
Another strategy is to apply matrix sketching. These can include fast random projection methods \cite{ailon2006approximate} as well as deterministic sketching methods \cite{liberty2013matrixsketching}.

When the influence function is on the predicted values rather than the parameters themselves, section \ref{sec:prediction influence} shows that the influence of a point can sometimes be expressed as the product of a function of the residual and the leverage of the point. Methods to approximately compute leverage scores 
\cite{drineas2012fast} may be applied. 
The simplest leverage score approximation is to assume equal leverage of all points. The resulting sampling weight depends only on the residual.

\section{Examples}
We provide two examples for the influence function
in addition to the linear least squares model that is already derived in 
section \ref{sec:asymptotically linear}.
First, we consider the common case of generalized linear model.
Second, we consider a non-differentiable quantile regression problem.

\subsection{Generalized Linear Model}
Generalized linear models (GLMs) are a generalization of the normal linear model to allow for non-normal error distributions such as when responses are discrete valued. They are flexible and commonly used. Logistic, Poisson, and exponential regression models are examples of GLMs. 

A generalized linear model with a canonical link function has log likelihood given by $\log p(y_i, x_i | \theta) = y_i x_i^T \theta - A(x_i^T \theta)$
where $A(t) = \log \int_\Omega exp(y t) dy$ is the log partition function and $\Omega$ is the set of possible $y$ values.
Under smoothness conditions on the error distribution and when the model is correctly specified, 
the influence function is given by the rescaled score 
given in equation \ref{eqn:mle influence}. 
A straightforward derivation gives that $A'(x_i^T\theta) = \E(Y_i | x_i^T\theta)$ and $A''(x_i^T\theta) = 1/\var (Y_i | x_i^T\theta)$.
Thus, the score function is
$s_{\theta}(x_i, y_i) = (y_i - \hat{y}_i) x_i$ and the Fisher information is
$(X^TWX)$ where $W$ is a diagonal weight matrix with 
$W_{ii} = 1/\var_{\theta} (Y_i | x_i^T \theta)$. 
Thus the influence functions for $\hat{\theta}$ and $\hat{y}$ under correct model specification are given by 
\begin{align}
\psi_{\theta}(x_i, y_i) &= (y_i - \hat{y}_i) (X^T W X)^{-1} x_i \\
\psi^{(pred)}_\theta(x_i, y_i) &=  (y_i - \hat{y}_i) W X (X^TWX)^{-1} x_i =
r_i H_{i \cdot}^T.
\end{align}
where $r_i$ is the residual and $H = X (X^TWX)^{-1} X^T W$ so that 
$\hat{\theta} = H Y$. 
Unlike linear least squares regression, the matrix $H$ is non-symmetric
so the norm of the influence function cannot be expressed exactly in terms of the diagonal of the hat matrix .

For the special case of logistic regression, local case-control sampling provides a sampling method that has both good empirical and theoretical properties. It chooses sampling probabilities proportional to the ``surprise''
$y_i(1-\hat{p}_i) + (1-y_i)\hat{p}_i$ so that a point is likely to be sampled only if it did not match the prediction. The surprise can also be expressed as the absolute value of the residual $|y_i - \hat{p}_i|$.
Thus, local case-control sampling is equivalent to influence-based sampling under the approximation that 
there is no essentially no effect due to the regression design. 
Alternatively, influence-based sampling can be seen as local case-control sampling but with the addition of information about the predictors or regression design $X$.

\subsection{Quantile regression}
Quantile regression \cite{koenker2005quantile} provides another useful generalization of linear models.
While linear least squares regression focuses on estimating the conditional mean, in some cases, the quantity of interest is not the average but the upper or lower tails of a distribution. For example, a charitable foundation may be interested in predicting conditional quantiles in order to set suggested donation amounts. 
In the case of median or $L_1$ regression, the true regression coefficients match the least squares coefficients when the error distribution is symmetric; however, median regression enjoys robustness properties that make it less sensitive to outliers.

For quantile regression, the loss function is the non-differentiable ``check'' function $\ell_\tau(x) = (1-\tau) x 1(x < 0) + \tau x 1(x \geq 0)$ rather than the squared residual.  
When the desired quantile is $\tau$ and the true conditional quantile is linear, the influence function is given by
$\psi(x_i, y_i) = [\tau(1-\tau)]^{-1} V^{-1} x\, \rho(y_i - x_i^T \theta)$
where $\rho$ is a subgradient of the loss, $\rho(z) = 1-\tau$ if $z < 0$ and $\tau$ if $z > 0$,
and $V = \int xx^T f(0 | x)dG(x)$ when the $X_i$ are randomly drawn from a distribution $X_i \sim G$ and the error $Y_i - x_i^T\theta$ has density $f(\cdot | x_i)$. 
In particular, if the error distribution is independent of the predictors $X$,
$\Sigma = \frac{1}{n} X^TX$ is a consistent estimator of $V$. This gives the following estimated influence functions on the coefficients and predictions 
\vspace{-0.1cm}
\begin{align}
\label{eqn:estimated quantile influence}
\hat{\psi}(x_i, y_i) &=  [\tau(1-\tau)]^{-1} \rho(r_i) \Sigma^{-1} x_i \\ 
\hat{\psi}^{(pred)}(x_i, y_i) &=  [\tau(1-\tau)]^{-1} \rho(r_i) H_{\cdot i}.\end{align}
We note that this influence function has the same form as the 
influence function for linear regression in equation \ref{eqn:linear influence function}. The residual $r_i$ in the influence function for linear regression is simply replaced by $\rho(r_i)$ in the quantile influence function. We will refer to $\rho(r_i)$ as the "residual" for quantile regression.

\section{Error analysis}

Let $\phi(\cdot)$ be some real-valued function on distributions in some P-Donsker class $\mathcal{F}$. Suppose it is Hadamard differentiable at $P_{\theta}$ under the uniform norm $\ell({\mathcal{F}})^{\infty}$
with influence function $\psi_\theta$. 
Assume values $X_i$ are drawn i.i.d.\@ from $P_{\theta}$.
Consider the set of measures $Q$ that are mutually absolutely continuous with respect to $P_{\theta}$ with $\alpha \leq dQ / dP \leq 1$ almost everywhere and the total measure of $Q$ is some constant $c$. 

$Q$ defines an importance sub-sampling distribution which is generated by taking the empirical distribution $\mathbb{P}_n$ and keeping item $x$ in the empirical distribution with probability $\pi_x = dQ/dP(x)$.
Let $\hat{\mathbb{P}}^Q_n = n^{-1} \sum_{i=1}^n Z_i \frac{dP}{dQ}(X_i) \delta_{X_i} $ be the resulting estimated empirical measure where $Z_i$ indicates $X_i$ is the the subsample and equals 1 with probability $\frac{dQ}{dP}$. 
Since the weights $\frac{dP}{dQ}$ are bounded and fixed, it follows that 
$\sqrt{\frac{n}{c}}(\hat{\mathbb{P}}^Q_n - P) \convd R^Q$ where $R^Q$ is tight in $\ell(\mathcal{F})^{\infty}$. The functional delta method \cite{vandervaart} gives that
$\sqrt{\frac{n}{c}}(\phi(\hat{\mathbb{P}}^Q_n) - \phi(P)) 
\convd \int \psi(x) dR(x)$.
Furthermore, this limit is $Normal(0,V^Q)$
where $V^Q = \int \psi(x)^2 \left(\frac{dP}{dQ}\right)^2 dQ(x)$.
It is thus sensible to define the optimal importance sub-sampling measure $Q_{opt}$ to be the one that minimizes the asymptotic variance $V^Q$.

Let $\hat{\pi} \in \Pi_{\alpha, n}$ be estimated regularized inclusion probabilities for a sample of expected size $m$ based on the influence function. These probabilities are asymptotically optimal in the sense that the resulting estimates converge to some limit Normal distribution where the variance of that limit distribution is equal to the limit variance under the optimal importance sub-sampling distribution. The proof is deferred to the supplementary material.

\begin{theorem}
	Suppose the pilot estimate of the influence function is consistent
	$\tilde{\psi} = \psi_{\theta_0} + o_p(1)$ under the uniform norm.
	As $m,n \to \infty$ with $m /n \to c > 0$, the plug-in estimator $\phi(\hat{\mathbb{P}}_n(\hat{\pi}) ) \convd Normal(0, V^{Q_{opt}})$.
\end{theorem}

\vspace{-0.1cm}
\section{Experiments}
\begin{figure}
\hspace{-0.5cm}\begin{tabular}{cc}
	Approximate & Exact \\
	\includegraphics[width=70mm]{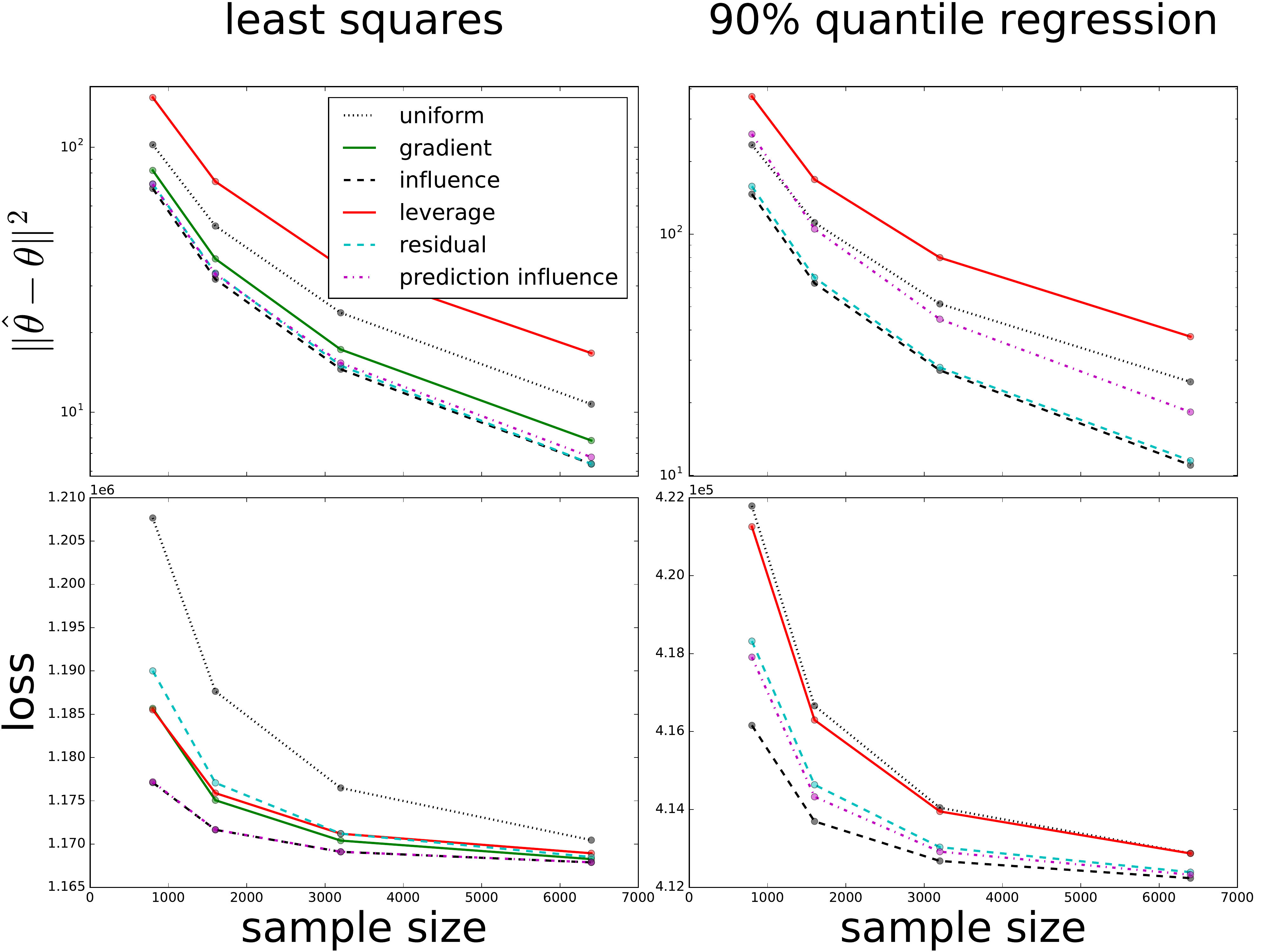}
&
\includegraphics[width=70mm]{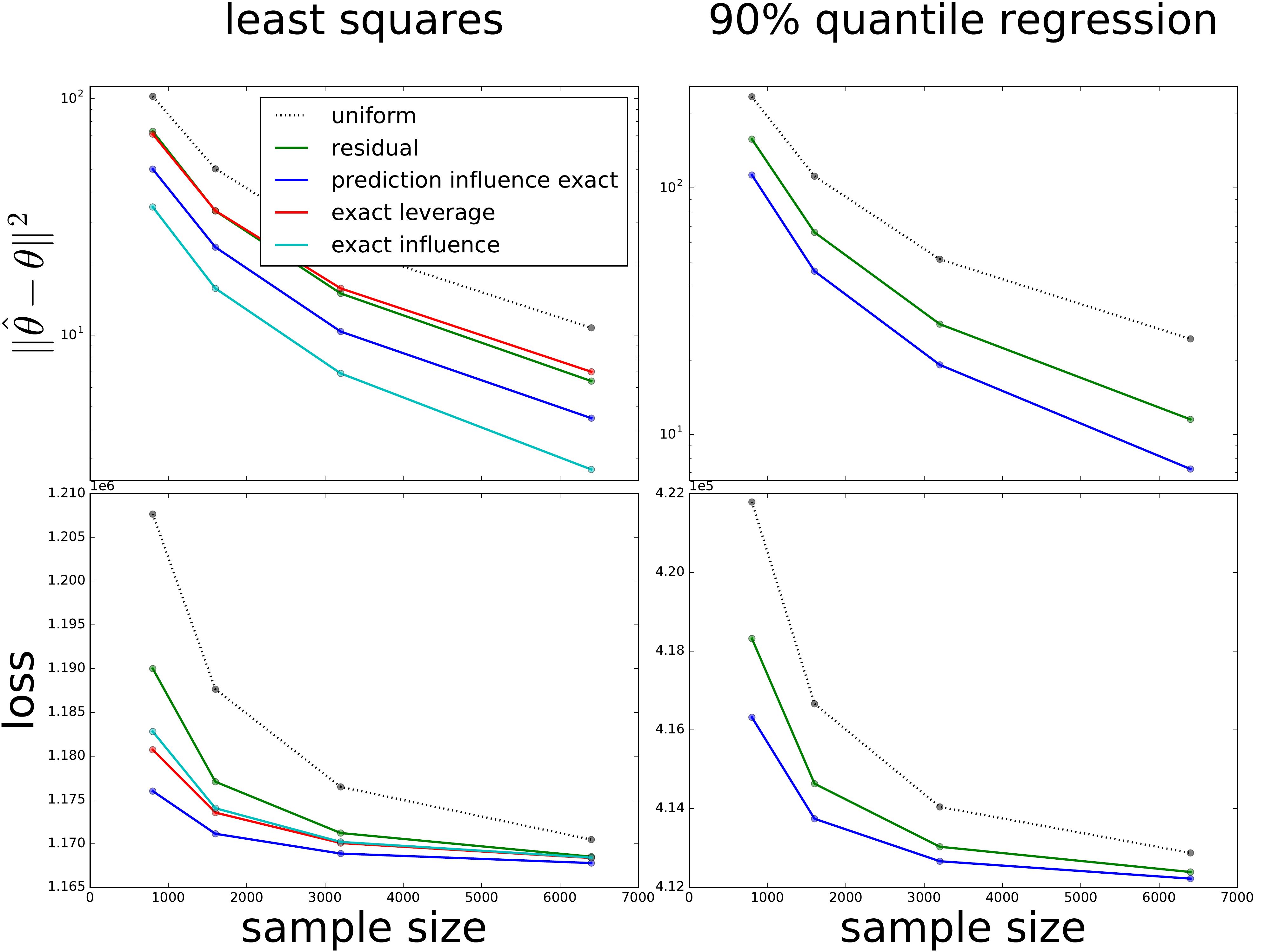}
\end{tabular}
\caption{Results of our method on the CASP data set \cite{Lichman:2013}. The left 4 subplots show the performance of methods based on approximations to the influence and leverage while the right 4 are based on the exact values. 	As expected, our influence-based methods are the top performers. They require $1/3$ to $1/2$ the sample size compared to uniform sampling to achieve the same error. This holds for both least squares and quantile regression tasks when their customized influence functions are used. We also find that the gradient-based sampling method performed worse than just using the residuals. This was likely due to the widely ranging scales of each variable as we did not make an additional pass to normalize the data. Furthermore, we note that the approximate-leverage-based method performs better than uniform sampling to minimize prediction error but worse in minimizing error in the coefficients.}
\label{fig:CASP}
\end{figure}

\begin{figure}
	\hspace{-0.5cm}\begin{tabular}{cc}
	\includegraphics[width=70mm]{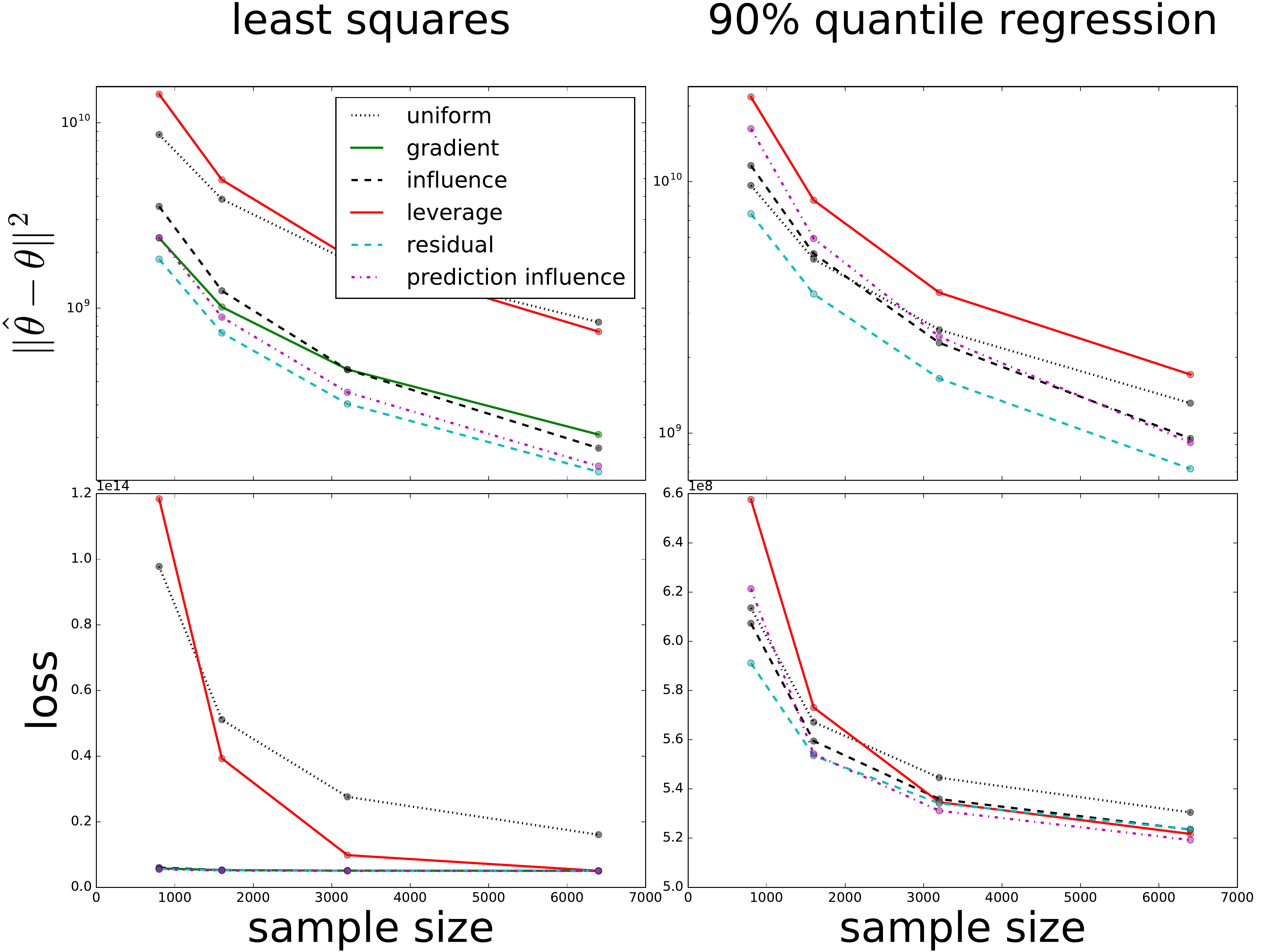}
&
	\includegraphics[width=70mm, height=60mm]{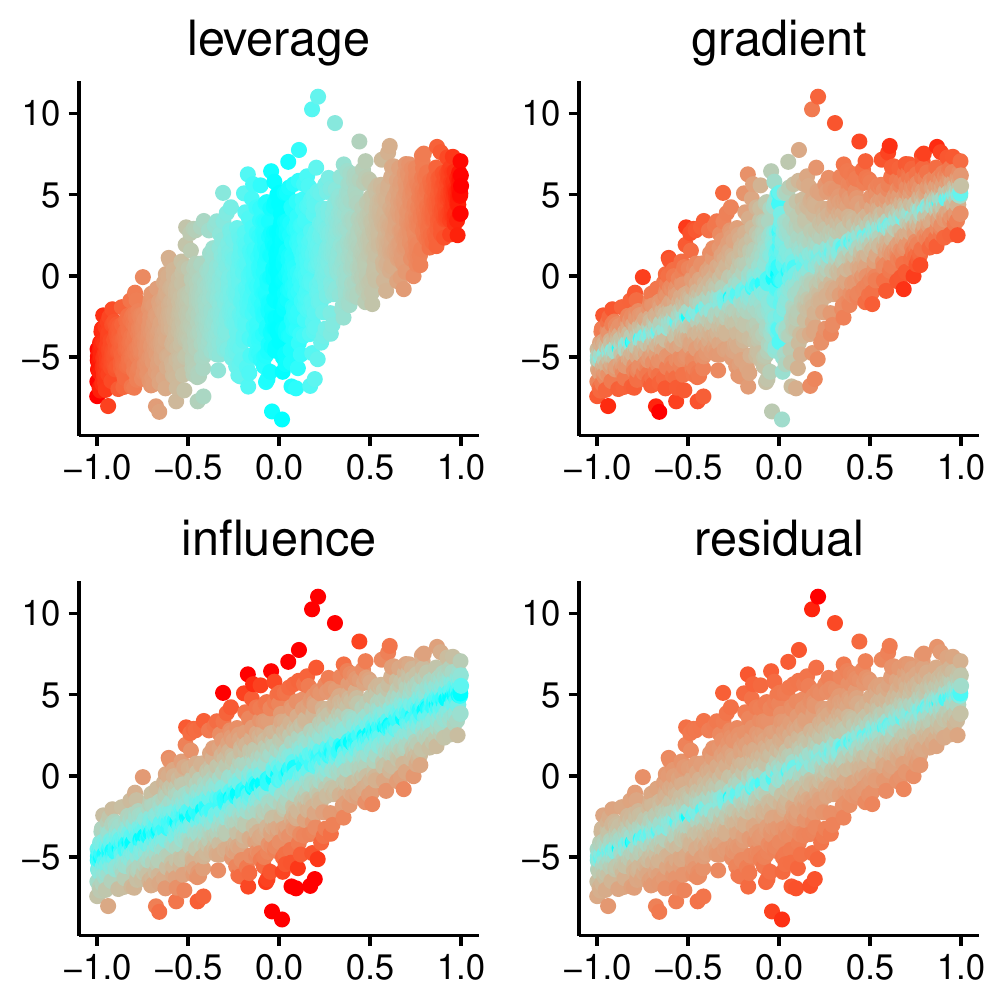}.
\end{tabular}
	\caption{{\bf Left:} Results of our method on the NEWS data set
	\cite{fernandes2015proactive}.  The results on the NEWS dataset tell a similar story to the CASP dataset but with much larger gains over uniform sampling for least squares regression. This is likely due to the heavy tailed nature of the NEWS dataset compared to CASP. The dataset also exhibits poor approximation of the matrix inverse needed to compute the ``influence'' of the design, so that simply using the residual with no approximated influence from the regression design performs best for least squares. {\bf Right:} Sampling patterns for different importance measures for the model $Y = 5 X + 1000 Z + \epsilon$ where $Z = 10^{-3}$. Red denotes high inclusion probability while light blue denotes low inclusion probability. The differences between the methods are made more apparent in the supplementary material.
	Leverage focuses only on the extremes of the design while completely ignoring large residual near the center. The gradient-based design also ignores large residuals near the center though to a much smaller degree. The influence-based design picks an appropriate balance of residual and regression design effects where points near the center are slightly less likely to be sampled given the same residual. The ``influence'' of the residual dominates over the ``influence'' of the regression design here.
	\label{fig:NEWS}
}
\end{figure}

We consider two real datasets from the UCI repository, the CASP \cite{Lichman:2013} ($n=45730$, $d=9$) and the Online News Popularity ($n=38644$, $d=59$) datasets, and show the results in Figures \ref{fig:CASP} and \ref{fig:NEWS}, respectively. We consider both least squares and quantile regression models.
For the Online News Popularity dataset, we removed 4 columns due to collinearity. In each case, we use a 5\% random sample of the data to derive a pilot estimate and drew a weighted sub-sample from the remainder.

We considered 2 linear regression models, linear least squares and quantile regression. The quality of the fit on a subsample is measured either by squared error in the coefficients $\|\hat{\theta} - \theta_{opt}\|^2$ 
or by the loss function corresponding to the optimization. As we wish to make only one pass through the data, we use the same approach as in \cite{dhillon2013new} for computing approximate matrix inverses needed for the influence and leverage rather than the fast Johnstone-Lindenstrauss transforms needed in \cite{drineas2012fast}.
We also regularize the $d\times d$ matrix inverse by taking $(V+\lambda I)^{-1}$ where $\lambda = Trace(V) / 10d$.
For completeness, we compute the exact leverage scores and show that they still perform worse.

\section{Conclusion}
We have demonstrated both theoretically and empirically that influence functions yield good and principled sub-sampling procedures for handling big data. They allow one to fundamentally understand and recast the problem as optimal sampling for a mean estimation problem where the mean is taken over influences. In particular, we show that our approach yields the best possible asymptotic variance over all Poisson sampling designs with the same size and regularization. Our approach can be applied to a wide range of statistical and machine learning models. Furthermore, although the full influence often requires a matrix inversion, simple approximations to the influence that take only $O(nd)$ time, such as using only the residual, can perform well while being easy to both compute and implement.

\clearpage
\bibliographystyle{abbrv}
\bibliography{dtingbib}

\appendix

\section{Proof}
The proof consists of two parts. One establishes that sampling with probability proportional to influence leads to the optimal importance sub-sampling measure. This argument is a straightforward application of the same argument in section \ref{sec:optimal design}.

The second consists of showing that the sample drawn using the estimated influences effectively yields the same sample as one using the exact influences. The consistency of the estimated influence $\hat{\psi}$
gives that $\hat{\pi} = dQ / dP + o_p(1)$. 
For Poisson sampling with , the item $X_i, Y_i$ is in the sample if $U_i < \pi_i$
for independent $U_i \sim Uniform(0,1)$. 
Let $\pi_i = dQ/dP(X_i)$ and $\tilde{\pi}_i$ be the estimated inclusion probability using the estimated influence.
Let $Z_i = 1$ if $U_i < \pi_i$ and $0$ otherwise. 
Likewise, $\tilde{Z}_i = 1$ if $U_i <\tilde{\pi}_i$ and 0 otherwise.
Let $\epsilon_i = \tilde{\pi}_i - \pi_i$.
The difference 
\begin{align}
Z_i / \pi_i - \tilde{Z}_i / \tilde{\pi}_i &=
\frac{Z_i - \tilde{Z}_i}{\pi_i} (1 - \epsilon_i + O(\epsilon_i^2))
\end{align}
Taking the numerator, we have $|Z_i - \tilde{Z}_i| \sim Bernoulli(\epsilon_i)$.
And the overall expectation of the absolute value is $O(\epsilon_i / \pi_i)$.
Since $\var (Z_i / \pi_i) = (1-\pi_i)$, it follows that 
the empirical estimate based on the estimated influences
$\sqrt{\frac{n}{c}}(\hat{\mathbb{P}} - P)$
converges to the same limit as that under the optimal $Q_{opt}$.
Hence, $\sqrt{m}(\phi(\hat{\mathbb{P}}) - \phi(P))$
and $\sqrt{m}(\phi(\hat{\mathbb{P}}^{Q_{opt}}) - \phi(P))$
also converge to the same limit by the functional delta method.

\section{Additional figures for NEWS dataset}
\begin{figure}[H]
	\centering
	\includegraphics[width=180mm]{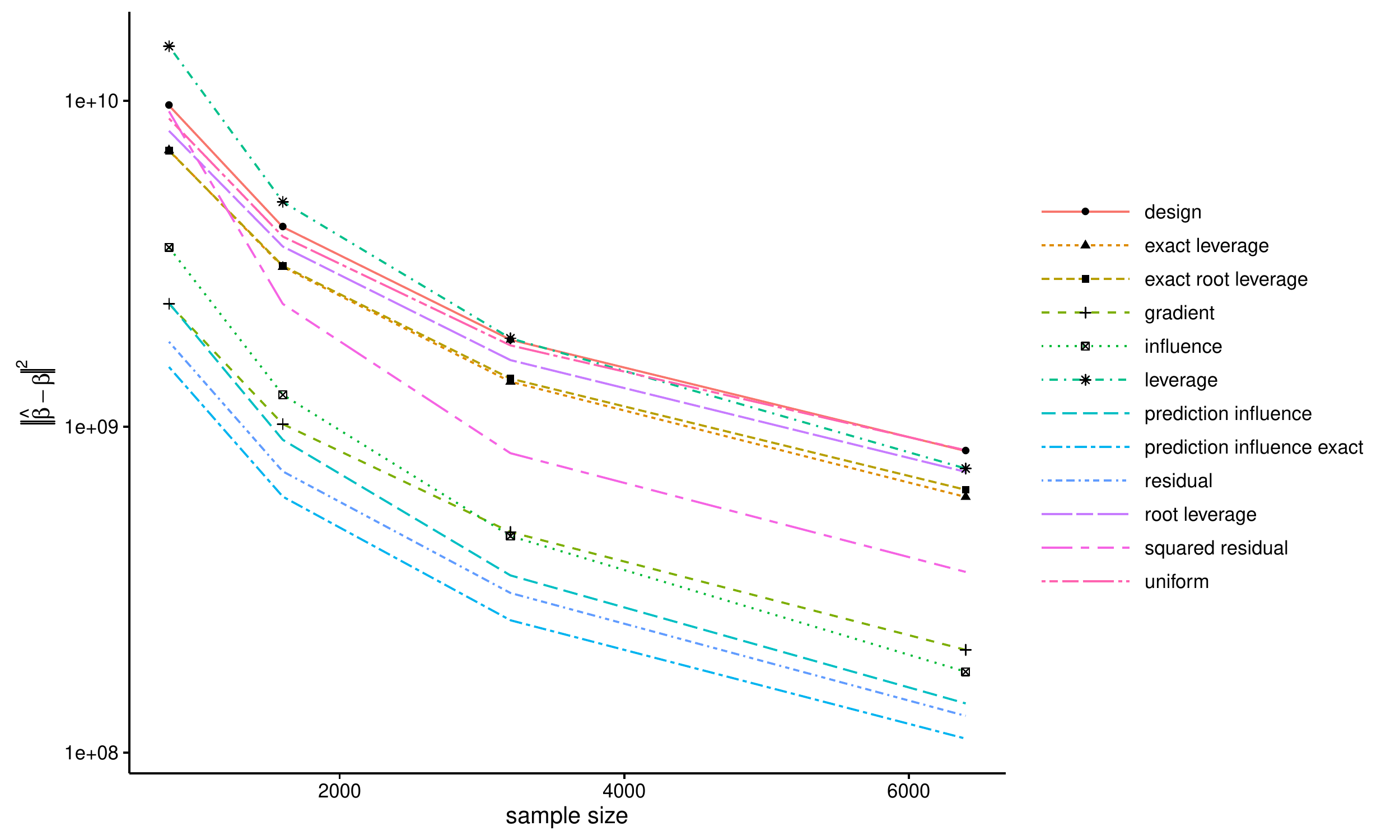}
	\label{fig:NEWSbeta}
\end{figure} 
\begin{figure}[H]
	\centering
	\includegraphics[width=180mm]{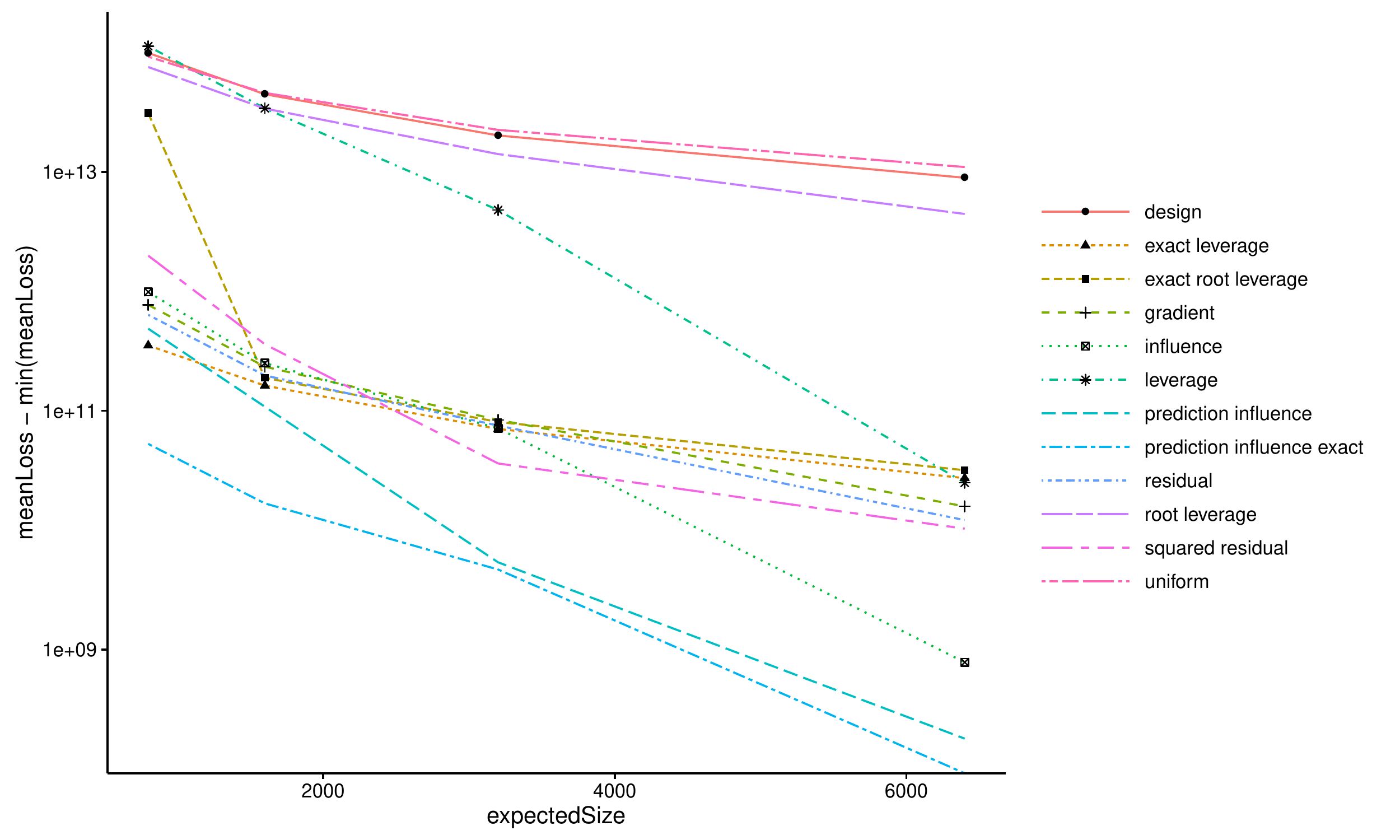}
	\label{fig:NEWSloss}
\end{figure} 

\clearpage

\section{Additional figure for CASP dataset}
\begin{figure}[H]
	\centering
	\includegraphics[width=180mm]{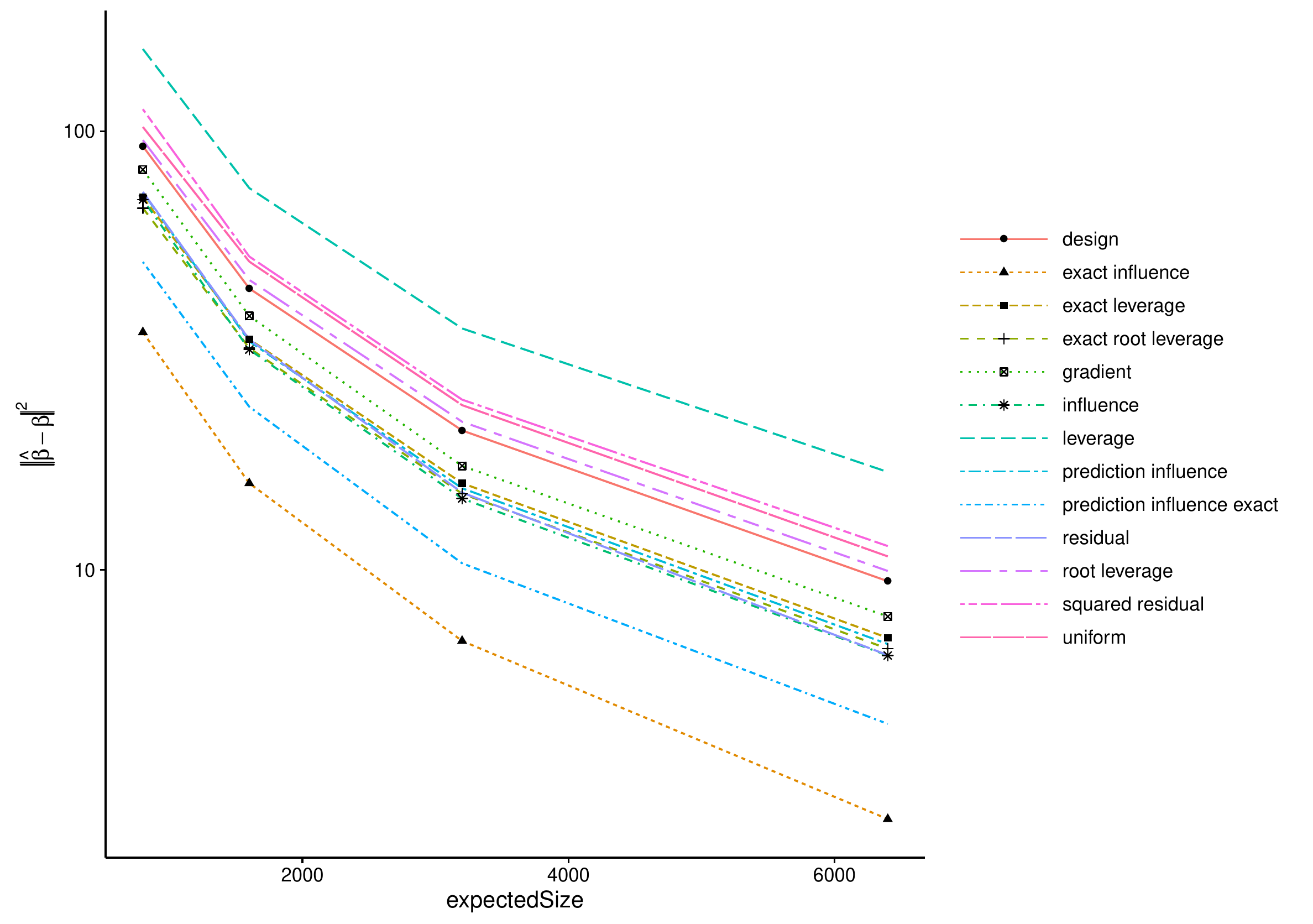}
	\label{fig:CASPbeta}
\end{figure} 
\begin{figure}[H]
	\centering
	\includegraphics[width=180mm]{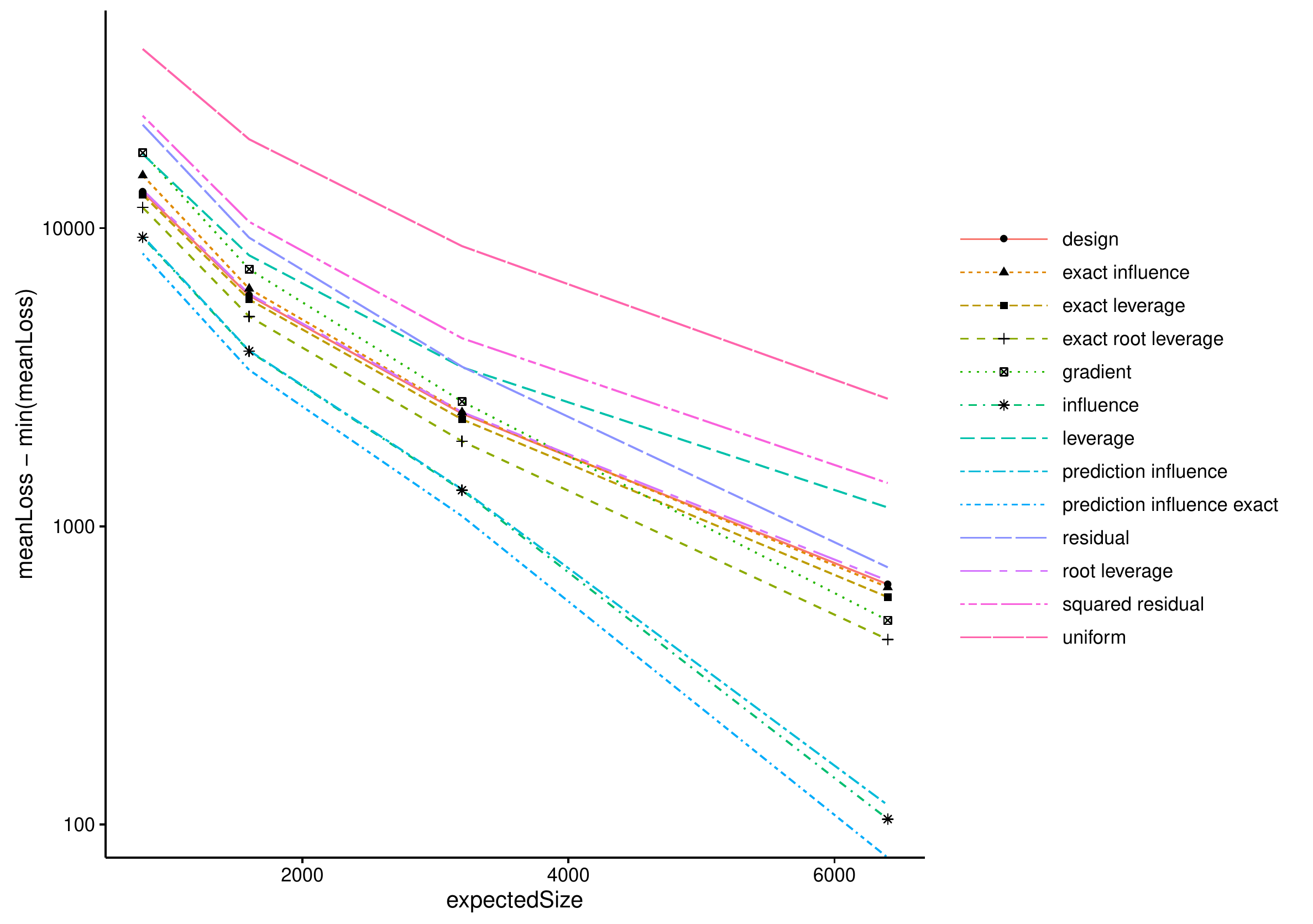}
	\label{fig:CASPloss}
\end{figure} 

\end{document}

%% file: macros.tex
\newcommand{\eat}[1]{}

\floatstyle{ruled}
\newfloat{algorithm}{tb}{loa}
\floatname{algorithm}{Algorithm}
\newtheorem{theorem}{Theorem}

\DeclareFontFamily{U}{mathx}{\hyphenchar\font45}
\DeclareFontShape{U}{mathx}{m}{n}{
      <5> <6> <7> <8> <9> <10>
      <10.95> <12> <14.4> <17.28> <20.74> <24.88>
      mathx10
      }{}
\DeclareSymbolFont{mathx}{U}{mathx}{m}{n}
\DeclareMathSymbol{\bigtimes}{1}{mathx}{"91}

\def\var{{\rm Var}}
\def\E{{\ensuremath{\mathbb E}}}

\def\convd{\rightsquigarrow} 
\long\def\comment#1{}

\renewcommand{\P}{\ensuremath{\mathbb P}}

